\documentclass{llncs}
\usepackage{amsmath,amssymb,epsf}
\usepackage{mathtools}
\usepackage{multirow}
\usepackage{url}

\begin{document}

\title{Multi-D Kneser-Ney Smoothing Preserving the Original Marginal Distributions\thanks{Published in: Dob\'o, Andr\'as (2017). Multi-D Kneser-Ney Smoothing Preserving the Original Marginal Distributions. Research in Computing Science, 141.}}

\author{
Andr\'as Dob\'o}
\institute{University of Szeged\\
          Institute of Informatics\\
          Szeged, Hungary\\
          dobo@inf.u-szeged.hu}

\maketitle

\begin{abstract}

Smoothing is an essential tool in many NLP tasks, therefore numerous techniques have been developed for this purpose in the past. One of the most widely used smoothing methods are the Kneser-Ney smoothing (KNS) and its variants, including the Modified Kneser-Ney smoothing (MKNS), which are widely considered to be among the best smoothing methods available. Although when creating the original KNS the intention of the authors was to develop such a smoothing method that preserves the marginal distributions of the original model, this property was not maintained when developing the MKNS.

In this article I would like to overcome this and propose such a refined version of the MKNS that preserves these marginal distributions while keeping the advantages of both previous versions. Beside its advantageous properties, this novel smoothing method is shown to achieve about the same results as the MKNS in a standard language modelling task.

\end{abstract}

\section{Introduction}

The goal of smoothing is to overcome data sparsity, which poses a huge problem in numerous tasks, including a vast number of NLP problems. A very good example of this is language modelling, where the task is to learn the probability of word sequences given some training data: using lower order models for this purpose do not provide sufficient context, while choosing large models will usually suffer from insufficient training data (for an n-gram model there are $|v|^{n}$ distinct n-gram types, where $|v|$ is the size of the vocabulary). Due to this, most of the values in a basic n-gram model are equal to zero, which produces zero probabilities for most word sequences when simply using maximum-likelihood estimation. To overcome this, smoothing techniques have been widely used since decades, decreasing the probability of seen events and redistributing the gained probability mass among unseen events so as to avoid zero probabilities during prediction.

Due to their importance, smoothing methods have received considerable attention in the past. One of the most widely used group of smoothing methods are of the type absolute discounting \cite{ney91}, that are simple but still very powerful and efficient methods. The Kneser-Ney smoothing (KNS) \cite{kneser95}, and its multi-discount variant, the Modified Kneser-Ney smoothing (MKNS) \cite{chen99} are widely considered to be one of the best smoothing algorithms since a long time \cite{chen99,goodman01,teh06,siivola07,sundermeyer11,zhang14}.

Although the probability of atomic events changes during smoothing as a necessary consequence, the marginal probabilities do not necessarily need to change, where the marginal probabilities are the probabilities obtained by summing out the probabilities of an event with respect to other events:

\begin{equation}
P(Y) = \sum_{z \in Z}P\left(Y,z\right)
\end{equation}

One of the key motivations when developing the KNS was that it should preserve the marginal distributions of the original model, meaning that the obtained model satisfies the following equation:

\begin{equation}
\frac{c\left(w_i\right)}{\sum_{w_i}c\left(w_i\right)} = \sum_{w_{i-1}}p\left(w_i|w_{i-1}\right)p\left(w_{i-1}\right)
\end{equation}

This is very advantageous in many cases, and under certain assumptions, an optimal model can only be obtained by satisfying this property, as discussed by Goodman in the extended version of his paper \cite{goodman01}. Hence Goodman comes to the conclusion that under these assumptions any smoothing method not preserving the original marginals can be improved by modifying it to preserve them. Despite this fact, many frequently used smoothing techniques, including the MKNS, do not satisfy this property: when Chen and Goodman \cite{chen99} refined the original KNS by introducing three discount parameters instead of just one, they did not adjust the lower-order distributions according to this change, which resulted in the loss of the original marginals in the smoothed model.

Within this article I present such a novel smoothing method based on the MKNS, that keeps all the advantages of both the KNS and the MKNS, while also preserving the original marginal distributions. Section 2 gives a general overview of smoothing methods and introduces the problem of preserving the marginal distributions in detail. The presentation of my novel method (called MDKNSPOMD) follows in Section 3, which is evaluated on a standard language modelling task in Section 4. Section 5 gives a short summary and draws conclusions.

\section{Background}

One of the earliest and simplest smoothing algorithms is called Add-k smoothing \cite{laplace25,chen99}. This basically adds a fixed constant value (often simply 1) to the count of each observed and unobserved event, and computes the probabilities on these modified counts. As in case of this model too, in the rest of the article I will present each formula and example adapted to bigram language modelling (when not noted otherwise). However, all methods can be used generally on higher-order models and on other applicable data types too. The bigram formula for the Add-k smoothing, with a smoothing parameter $\delta$ and a vocabulary size of $|V|$, is as follows:

\begin{equation}
p_{Add}\left(w_i|w_{i-1}\right) = \frac{\delta + c\left(w_{i-1}w_i\right)}{\delta |V| + \sum_{w_i}c\left(w_{i-1}w_i\right)}
\end{equation}

Due to its simplicity, this method is used very widely and can actually work very well in some cases, especially if there are not too many zero counts. However, in case of most problems, especially with huge models full of zeros (such as language modelling), it overweighs unseen events and does not work too well. Besides, it also does not keep the original marginals.

Another frequently used method is the Good-Turing smoothing (GT) \cite{good53}, trying to estimate the probability of words that occurred r times using the frequency of words seen (r+1) times:

\begin{equation}
r^* = \left(r+1\right)\frac{n_{r+1}}{n_r}
\end{equation}

, where $n_r$ is the number of n-grams that are present in the corpora exactly $r$ times. Based on this the probability of a bigram occurring $r$ times is given as: 

\begin{equation}
p_{GT}\left(w_i|w_{i-1}\right) = \frac{r^*}{\sum_{w_i}c\left(w_i\right)}
\end{equation}

Although this method has a very good intuition and can sometimes work quite well, it does not combine higher-order models with lower-order models, and just uses the same general smoothing for all words with count $r$. Therefore this is usually outperformed by more sophisticated methods.

A very popular category of smoothing methods try to estimate the probabilities in an n-gram model by also making use of information from an (n-1)-gram model. This is advantageous as there are much less (n-1)-gram types then n-gram types, so the number of zeros in the (n-1)-gram model is much less then in case of the n-gram model. These techniques either back off to the lower order model or interpolate the higher-order model with it. As interpolation is fairly consistently more successful than backing off \cite{chen99,goodman01}, in the rest of the article I will only consider this version of each smoothing algorithm. Those smoothing methods that back off from higher-order models or interpolate them with lower order models can be recursively applied to the lower order models too, which further helps eliminating the zero probabilities and usually helps to achieve better results. 

One of the easiest ways to create an interpolated model is by simple absolute discounting (Abs) \cite{ney91}. The motivation behind this was that looking at the results of other smoothing methods, such as the Good-Turing smoothing, one can often notice that it is as if the same value was simply subtracted from the count of each seen event ($D<1$), hence it would be easier to just simply do this instead of doing more complex calculations: 

\begin{equation}
p_{Abs}\left(w_i|w_{i-1}\right) = \frac{c\left(w_{i-1}w_i\right) - D}{\sum_{w_i}c\left(w_{i-1}w_i\right)} +\left(1-\lambda_{w_{i-1}}\right)p_{Abs}\left(w_i\right)
\end{equation}

Despite its simplicity, absolute discounting can work quite well, and it was the basis for many of the most successful discounting methods currently in use. Among these techniques are the Witten-Bell \cite{witten91}, the Jelinek-Mercer \cite{jelinek80} and the Kneser-Ney smoothing (KNS) \cite{kneser95}, and their variants.

The motivation behind the original KNS was to implement absolute discounting in such a way that would keep the original marginals unchanged, hence preserving all the marginals of the unsmoothed model. Their model is as follows (actually, the original article \cite{kneser95} only presented a backoff version, and the interpolated version shown here was only introduced by \cite{chen99}):

\begin{equation}
p_{KNS}\left(w_i|w_{i-1}\right) = \frac{c\left(w_{i-1}w_i\right) - D}{\sum_{w_i}c\left(w_{i-1}w_i\right)} +\gamma_{KNS}\left(w_{i-1}\right)p_{KNS}\left(w_i\right)
\end{equation}

The $\gamma_{KNS}\left(w_{i-1}\right)$ serves as normalization and should be chosen in a way so that the distributions of the words sum up to 1. For that the sum of the $\gamma_{KNS}\left(w_{i-1}\right)$ weights for a word should be the same as the sum of the discounts subtracted from the probabilities of the word:

\begin{equation}
\gamma_{KNS}\left(w_{i-1}\right) = \frac{N_{1+}\left(w_{i-1}.\right)D}{\sum_{w_i}c\left(w_{i-1}w_i\right)}
\end{equation}

with the $N$ functions defined as follows:

\begin{equation}
\begin{aligned}
N_{c}\left(w_{i-1}.\right) &= |\{w_{i} : c\left(w_{i-1}w_{i}\right)=c\}| \\
N_{c+}\left(w_{i-1}.\right) &= |\{w_{i} : c\left(w_{i-1}w_{i}\right) \geq c\}| \\
N_{c}\left(.w_{i}\right) &= |\{w_{i-1} : c\left(w_{i-1}w_{i}\right)=c\}| \\
N_{c+}\left(.w_{i}\right) &= |\{w_{i-1} : c\left(w_{i-1}w_{i}\right) \geq c\}| \\
N_{c}\left(..\right) &= |\{w_{i-1},w_{i} : c\left(w_{i-1}w_{i}\right)=c\}| \\
N_{c+}\left(..\right) &= |\{w_{i-1},w_{i} : c\left(w_{i-1}w_{i}\right) \geq c\}| \\
\end{aligned}
\end{equation}

In this model the discount parameter and the lower-order distribution are the free parameters, as the count values are given by the training data and the normalization is given based on the other parameters. Therefore one can implement different versions of this model by changing these two free parameters.
By choosing the right lower-order distribution it is possible to preserve the marginal probabilities. To achieve this one has to define the lower order distribution to return a probability of $p$ for a word $w_{i}$, where $p$ is the proportion of the discounts subtracted from the $c(.w_i)$ counts as compared to the discounts subtracted from all the $c(..)$ values:

\begin{equation}
p_{KNS}\left(w_i\right) = \frac{ N_{1+}\left(.w_i\right) } { N_{1+}\left(..\right) }
\end{equation}

However, please note that the marginals are only preserved in case of bigram models or in case of such higher-order models where the highest order model is simply interpolated with the unsmoothed second-to-highest order model. In case the second-to-highest order model is smoothed recursively the same way, then this property of the KNS is lost.

Chen and Goodman \cite{chen99} introduced an improved version of the original KNS by changing one of its free parameters, namely the discount parameter. They showed that the optimal discount values for counts of 1 and 2 are very different from the optimal discount value for higher counts. Therefore they proposed to have three discounting parameters ($D_1<1$, $D_2<2$ and $D_{3+}<3$) instead of just one, for counts of 1, 2 and at least 3, respectively:

\begin{equation} \label{eq:MKNS}
\begin{aligned}
p_{MKNS}\left(w_i|w_{i-1}\right) = &\frac{c\left(w_{i-1}w_i\right) - D\left(c\left(w_{i-1}w_i\right)\right)}{\sum_{w_i}c\left(w_{i-1}w_i\right)} + \\
&\gamma_{MKNS}\left(w_{i-1}\right)p_{MKNS}\left(w_i\right)
\end{aligned}
\end{equation}

where

\begin{equation} \label{eq1} 
D\left(x\right)=
\begin{dcases}
0 & if \enspace x=0\\
D_1 & if \enspace x=1\\
D_2 & if \enspace x=2\\
D_{3+} & if \enspace x\geq3
\end{dcases} 
\end{equation} 

With this modification, the normalization factor, in order to have all the word distributions sum up to 1, should be defined as follows:

\begin{equation} \label{eq:gammaMKNS}
\gamma_{MKNS}\left(w_{i-1}\right) = \frac{D_1N_1\left(w_{i-1}.\right) + D_2N_2\left(w_{i-1}.\right) + D_{3+}N_{3+}\left(w_{i-1}.\right)}{\sum_{w_i}c\left(w_{i-1}w_i\right)}
\end{equation}

As also noted earlier, the unigram distribution remains the same as it was in case of the KNS:

\begin{equation}
p_{MKNS}\left(w_i\right) = p_{KNS}\left(w_i\right)
\end{equation}

The optimal discount parameters for the KNS and MKNS models can be estimated as follows \cite{chen99}: 

\begin{equation}
\begin{aligned}
D &= \frac{n_1}{n_1+2n_2} \\[4pt]
D_1 &= 1-2D\frac{n_2}{n_1} \\[4pt]
D_2 &= 2-3D\frac{n_3}{n_2} \\[4pt]
D_{3+} &= 3-4D\frac{n_4}{n_3} \\[4pt]
\end{aligned}
\end{equation}

where $n_r$ represents the total number of n-grams with a frequency of $r$.

Although the discounting method in the MKNS is different then in the KNS, the authors left the calculation of the lower-order distributions unchanged, which results in not preserving the original marginal distributions. There already exist a couple of studies discussing this issue. Some simply note this fact \cite{teh06,siivola07,sundermeyer11}, without presenting any detailed discussion about it, while others also get into more detail.

For example, Zhang and Chiang \cite{zhang14} also note that this property of the MKNS can be resolved, but they do not provide a solution for this. Further, Chen and Rosenfeld \cite{chen00} note that using maximum entropy (ME) techniques one can obtain such models that preserve the original marginals. However, they do not apply this to the MKNS and only discuss in detail a Fuzzy ME model that only approximately preserves them. Although Roark et al. \cite{roark13} presents such a method that takes an arbitrary backoff smoothing model and transforms it into a model that preserves the original marginals, they do not test this technique on the MKNS model. So despite the earlier studies considering this problem, to my best knowledge, my study is the first to derive the solution for the MKNS and to perform thorough tests comparing the original KNS and MKNS methods with this new method. Therefore this study is novel in this respect.

To help better understand the difference between the smoothing methods and to give an easy insight into what preserving or not preserving the marginals means, I hereby present the joint and marginal counts of a sample bigram maximum likelihood model, unsmoothed and smoothed with KNS and MKNS, trained on the same small text in Tables 1, 2 and 3, respectively. In case of every table, each row corresponds to a value of x, each column represents a value of y, with the cells containing the c(x,y) values. The \textless{}s\textgreater{} and \textless{}/s\textgreater{} symbols are special symbols representing the beginning and end of the sentences, respectively. Inside the tables counts are presented instead of probabilities, as this way it is much easier to see whether the original marginals are preserved after smoothing or not. The sum of the counts in each column (which corresponds to the respective marginal) are presented at the end of them.

\begin{table}[p]
\caption{Joint and marginal counts of a simple maximum likelihood model trained on the sample text.}
\label{tab:table1}
\begin{center}
\begin{tabular}{|c|rrrrrrr|r|}
\hline
c(x, y) & \hspace{1pt} \textless s\textgreater & a & b & c & d & e & \hspace{1pt} \textless /s\textgreater \hspace{1pt} & \\
\hline
\textless s\textgreater & 0 & 2 & 3 & 5 & 0 & 1 & 0 \hspace{1pt} & 11 \\
a & 0 & 4 & 1 & 4 & 3 & 8 & 1 \hspace{1pt} & 21 \\
b & 0 & 7 & 2 & 1 & 0 & 0 & 4 \hspace{1pt} & 14 \\
c & 0 & 2 & 5 & 2 & 0 & 4 & 2 \hspace{1pt} & 15 \\
d & 0 & 1 & 0 & 0 & 2 & 0 & 3 \hspace{1pt} & 6 \\
e & 0 & 5 & 3 & 3 & 1 & 6 & 1 \hspace{1pt} & 19 \\
\textless /s\textgreater & 0 & 0 & 0 & 0 & 0 & 0 & 0 \hspace{1pt} & 0 \\
\hline
 & 0 & 21 & 14 & 15 & \hspace{2pt} 6 & 19 & 11 \hspace{1pt} & 86 \\

\hline

\end{tabular}
\end{center}
\end{table}

\begin{table}[p]
\caption{Joint and marginal counts of a simple maximum likelihood model with KNS trained on the sample text.}
\label{tab:table2}
\begin{center}
\begin{tabular}{|c|rrrrrrr|r|}
\hline
c(x, y) & \hspace{1pt} \textless s\textgreater & a & b & c & d & e & \hspace{1pt} \textless /s\textgreater \hspace{1pt} & \\
\hline
\textless s\textgreater & 0.00 & 1.95 & 2.89 & 4.89 & 0.16 & 0.84 & 0.26 \hspace{1pt} & 11.00 \\
a & 0.00 & 4.11 & 1.03 & 4.03 & 2.87 & 7.95 & 1.03 \hspace{1pt} & 21.00 \\
b & 0.00 & 6.95 & 1.89 & 0.89 & 0.16 & 0.21 & 3.89 \hspace{1pt} & 14.00 \\
c & 0.00 & 2.03 & 4.96 & 1.96 & 0.20 & 3.89 & 1.96 \hspace{1pt} & 15.00 \\
d & 0.00 & 0.87 & 0.20 & 0.20 & 1.75 & 0.16 & 2.83 \hspace{1pt} & 6.00 \\
e & 0.00 & 5.11 & 3.03 & 3.03 & 0.87 & 5.95 & 1.03 \hspace{1pt} & 19.00 \\
\textless /s\textgreater & 0.00 & 0.00 & 0.00 & 0.00 & 0.00 & 0.00 & 0.00 \hspace{1pt} & 0.00 \\
\hline
 & 0.00 & 21.00 & 14.00 & 15.00 & \hspace{2pt} 6.00 & 19.00 & 11.00 \hspace{1pt} & 86.00 \\
\hline

\end{tabular}
\end{center}
\end{table}

\begin{table}[p]
\caption{Joint and marginal counts of a simple maximum likelihood model with MKNS trained on the sample text.}
\label{tab:table3}
\begin{center}
\begin{tabular}{|c|rrrrrrr|r|}
\hline
c(x, y) & \hspace{1pt} \textless s\textgreater & a & b & c & d & e & \hspace{1pt} \textless /s\textgreater \hspace{1pt} & \\
\hline
\textless s\textgreater & 0.00 & 2.01 & 2.09 & 4.09 & 0.55 & 1.36 & 0.91 \hspace{1pt} & 11.00 \\
a & 0.00 & 3.90 & 2.06 & 3.61 & 2.04 & 7.32 & 2.06 \hspace{1pt} & 21.00 \\
b & 0.00 & 6.27 & 1.83 & 1.54 & 0.55 & 0.73 & 3.09 \hspace{1pt} & 14.00 \\
c & 0.00 & 2.40 & 4.41 & 2.15 & 0.74 & 3.16 & 2.15 \hspace{1pt} & 15.00 \\
d & 0.00 & 1.33 & 0.58 & 0.58 & 1.27 & 0.47 & 1.76 \hspace{1pt} & 6.00 \\
e & 0.00 & 4.90 & 2.61 & 2.61 & 1.49 & 5.32 & 2.06 \hspace{1pt} & 19.00 \\
\textless /s\textgreater & 0.00 & 0.00 & 0.00 & 0.00 & 0.00 & 0.00 & 0.00 \hspace{1pt} & 0.00 \\
\hline
 & 0.00 & 20.80 & 13.58 & 14.58 & \hspace{2pt} 6.63 & 18.36 & 12.04 \hspace{1pt} & 86.00 \\
\hline

\end{tabular}
\end{center}
\end{table}

\section{My novel  method}

As previously presented, the MKNS is widely considered to be one of the best smoothing algorithms. However, despite the original motivation for its base variant (KNS), it does not have the property of keeping the original marginals unchanged. My novel method, which is called Multi-D Kneser-Ney Smoothing Preserving the Original Marginal Distributions (MDKNSPOMD), was developed to overcome this problem.

Its basis comes from the MKNS, with the idea for maintaining the marginal distributions from the original KNS, by changing one of the free parameters of the MKNS, namely the lower-order distribution. It is easy to see that the marginal distributions in a bigram model can be preserved if the lower order distribution is designed in a way that it returns a probability of $p$ for a word $w_{i}$, where $p$ is the proportion of the discounts subtracted from the count of the bigrams ending with $w_i$ as compared to all the discounts subtracted at the whole bigram level. This can easily be derived mathematically for the MKNS, in a similar manner as Chen and Goodman \cite{chen99} did it for the KNS. The derivation starts with the following equation:

\begin{equation}
\frac{c\left(w_i\right)}{\sum_{w_i}c\left(w_i\right)} = \sum_{w_{i-1}}p\left(w_i|w_{i-1}\right)p\left(w_{i-1}\right)
\end{equation}

, in which it is possible to express $p\left(w_{i-1}\right)$ by its empirical estimation from the training data:

\begin{equation}
p\left(w_{i-1}\right) = \frac{c\left(w_{i-1}\right)}{\sum_{w_{i-1}}c\left(w_{i-1}\right)}
\end{equation}

to get (after some simplification):

\begin{equation}
c\left(w_i\right) = \sum_{w_{i-1}}c\left(w_{i-1}\right)p\left(w_i|w_{i-1}\right)
\end{equation}

Then $p\left(w_i|w_{i-1}\right)$ can be substituted with its formula in MKNS (Equation \ref{eq:MKNS}):

\begin{equation}
c\left(w_i\right) = \sum_{w_{i-1}}c\left(w_{i-1}\right)\left(\frac{c\left(w_{i-1}w_i\right) - D\left(c\left(w_{i-1}w_i\right)\right)}{\sum_{w_i}c\left(w_{i-1}w_i\right)} +\gamma\left(w_{i-1}\right)p\left(w_i\right)\right)
\end{equation}

, after which $\gamma\left(w_{i-1}\right)$ can also be expressed as it is for MKNS in Equation \ref{eq:gammaMKNS}, and a couple of simplifying steps can be made to obtain:

\begin{equation}
\begin{aligned}
c\left(w_i\right) = \sum_{w_{i-1}} & c\left(w_{i-1}\right)\Bigg[ \frac{c\left(w_{i-1}w_i\right) - D\left(c\left(w_{i-1}w_i\right)\right)}{c\left(w_{i-1}\right)} + \\
                                                        & \frac{D_1N_1\left(w_{i-1}.\right) + D_2N_2\left(w_{i-1}.\right) + D_{3+}N_{3+}\left(w_{i-1}.\right)}{c\left(w_{i-1}\right)}p\left(w_i\right) \Bigg]
\end{aligned}
\end{equation}

\begin{equation}
\begin{aligned}
c\left(w_i\right) = \sum_{w_{i-1}} & \Big[ c\left(w_{i-1}w_i\right) - D\left(c\left(w_{i-1}w_i\right)\right) + \\
                                                        & \left(D_1N_1\left(w_{i-1}.\right) + D_2N_2\left(w_{i-1}.\right) + D_{3+}N_{3+}\left(w_{i-1}.\right)\right)p\left(w_i\right) \Big]
\end{aligned}
\end{equation}

\begin{equation}
\begin{aligned}
c\left(w_i\right) = &c\left(w_i\right) - \bigg( \sum_{w_{i-1}} D\left(c\left(w_{i-1}w_i\right)\right)\bigg) + \\
                               & p\left(w_i\right) \sum_{w_{i-1}} \left(D_1N_1\left(w_{i-1}.\right) + D_2N_2\left(w_{i-1}.\right) + D_{3+}N_{3+}\left(w_{i-1}.\right)\right)
\end{aligned}
\end{equation}

From here $p\left(w_i\right)$ can be easily expressed as: 

\begin{equation}
p\left(w_i\right) =  \frac{ \sum_{w_{i-1}}  D\left(c\left(w_{i-1}w_i\right)\right) } 
                                       { \sum_{w_{i-1}} \left(D_1N_1\left(w_{i-1}.\right) + D_2N_2\left(w_{i-1}.\right) + D_{3+}N_{3+}\left(w_{i-1}.\right)\right) }
\end{equation}

, and the sums can be rewritten to get the following form:

\begin{equation}
p\left(w_i\right) =  \frac{ D_1N_1\left(.w_{i}\right) + D_2N_2\left(.w_{i}\right) + D_{3+}N_{3+}\left(.w_{i}\right) } 
                                       { D_1N_1\left(..\right) + D_2N_2\left(..\right) + D_{3+}N_{3+}\left(..\right) }
\end{equation}

This proves that the MKNS smoothing model with the modification of using the above lower-order distribution will preserve the original marginal probabilities when interpolating with the above unigram distribution, and there are no other ways to achieve this. So this gives the following final form for the MDKNSPOMD for a bigram language model:

\begin{equation}
\begin{aligned}
p_{MDKNSPOMD}\left(w_i|w_{i-1}\right) = & \frac{c\left(w_{i-1}w_i\right) - D\left(c\left(w_{i-1}w_i\right)\right)}{\sum_{w_i}c\left(w_{i-1}w_i\right)} + \\
                                                                 &\gamma_{MDKNSPOMD}\left(w_{i-1}\right)p_{MDKNSPOMD}\left(w_i\right)
\end{aligned}
\end{equation}

\begin{equation}
\begin{aligned}
\gamma_{MDKNSPOMD}\left(w_{i-1}\right) = &\frac{D_1N_1\left(w_{i-1}.\right) + D_2N_2\left(w_{i-1}.\right)}{\sum_{w_i}c\left(w_{i-1}w_i\right)} + \\
& \frac{D_{3+}N_{3+}\left(w_{i-1}.\right)}{\sum_{w_i}c\left(w_{i-1}w_i\right)}
\end{aligned}
\end{equation}

\begin{equation}
p_{MDKNSPOMD}\left(w_i\right) = \frac{ D_1N_1\left(.w_{i}\right) + D_2N_2\left(.w_{i}\right) + D_{3+}N_{3+}\left(.w_{i}\right) } { D_1N_1\left(..\right) + D_2N_2\left(..\right) + D_{3+}N_{3+}\left(..\right) }			  
\end{equation}

To be able to easily see the working of this method, compare it with the KNS and MKNS models and to see its property of preserving the marginals of the original model, in Table 4 I present the joint and marginal counts of a sample bigram maximum likelihood model, smoothed with MDKNSPOMD, trained on the same small text as used in Tables 1, 2 and 3.

\begin{table}
\caption{Joint and marginal counts of a simple maximum likelihood model with \hspace{\textwidth} MDKNSPOMD trained on the sample text.}
\label{tab:table4}
\begin{center}
\begin{tabular}{|c|rrrrrrr|r|}
\hline
c(x, y) & \hspace{1pt} \textless s\textgreater & a & b & c & d & e & \hspace{1pt} \textless /s\textgreater \hspace{1pt} & \\
\hline
\textless s\textgreater & 0.00 & 2.04 & 2.15 & 4.15 & 0.46 & 1.45 & 0.76 \hspace{1pt} & 11.00 \\
a & 0.00 & 3.94 & 2.16 & 3.70 & 1.90 & 7.47 & 1.84 \hspace{1pt} & 21.00 \\
b & 0.00 & 6.30 & 1.89 & 1.60 & 0.46 & 0.82 & 2.94 \hspace{1pt} & 14.00 \\
c & 0.00 & 2.43 & 4.49 & 2.23 & 0.62 & 3.28 & 1.95 \hspace{1pt} & 15.00 \\
d & 0.00 & 1.35 & 0.62 & 0.62 & 1.21 & 0.52 & 1.67 \hspace{1pt} & 6.00 \\
e & 0.00 & 4.94 & 2.70 & 2.70 & 1.35 & 5.47 & 1.84 \hspace{1pt} & 19.00 \\
\textless /s\textgreater & 0.00 & 0.00 & 0.00 & 0.00 & 0.00 & 0.00 & 0.00 \hspace{1pt} & 0.00 \\
\hline
 & 0.00 & 21.00 & 14.00 & 15.00 & \hspace{2pt} 6.00 & 19.00 & 11.00 \hspace{1pt} & 86.00 \\
\hline

\end{tabular}
\end{center}
\end{table}

The smoothing method presented above for a bigram model can be generalized to higher-order models without a problem. To do this one has two possibilities. First, one can simply use the above model on the highest level of the model and interpolate it only with the second-to-highest level, not smoothing that level further. This version has the advantage that all the original marginals are preserved. However, one has to note that in case there are many zeros in the original model (such as in n-gram language models with $n\geq3$), this solution will not work well as it will still leave too many zeros in the model.

The other solution, as also proposed for other smoothing techniques in the past, is to do the interpolation recursively, always interpolating the $n^{th}$ level with the $(n-1)^{th}$ level, all the way back to the $1^{st}$ or $0^{th}$ level. In this case the theoretical and mathematical derivation previously applied on the bigram model for defining the right probability distributions in the lower level, can be used for all the levels. This would result in the following formula for a trigram model at the trigram level:

\begin{equation}
\begin{aligned}
p_{MDKNSPOMD}&\left(w_i|w_{i-2}w_{i-1}\right)  = \frac{c\left(w_{i-2}w_{i-1}w_i\right) - D_3\left(c\left(w_{i-2}w_{i-1}w_i\right)\right)}{\sum_{w_i}c\left(w_{i-2}w_{i-1}w_i\right)} \\
                                                           &+\gamma_{MDKNSPOMD}\left(w_{i-2}w_{i-1}\right)p_{MDKNSPOMD}\left(w_i|w_{i-1}\right)
\end{aligned}
\end{equation}

with $D_3$ being a discount function similar to the original $D$ discount function, with values $0$, $D_{(3;1)}$, $D_{(3;2)}$ and $D_{(3;3+)}$, and the normalization factor being:

\begin{equation}
\begin{aligned}
\gamma_{MDKNSPOMD}&\left(w_{i-2}w_{i-1}\right) = \frac{D_{(3;1)}N_1\left(w_{i-2}w_{i-1}.\right)}{\sum_{w_i}c\left(w_{i-2}w_{i-1}w_i\right)} + \\
                                                           & \frac{D_{(3;2)}N_2\left(w_{i-2}w_{i-1}.\right) + D_{(3;3+)}N_{3+}\left(w_{i-2}w_{i-1}.\right)}{\sum_{w_i}c\left(w_{i-2}w_{i-1}w_i\right)}
\end{aligned}
\end{equation}

With the same logic one gets the bigram level:

\begin{equation}
\begin{aligned}
&p_{MDKNSPOMD}\left(w_i|w_{i-1}\right) = \\
		  & \frac{ D_{(3;1)}N_1\left(.w_{i-1}w_{i}\right) + D_{(3;2)}N_2\left(.w_{i-1}w_{i}\right) } 
                                                                          { D_{(3;1)}N_1\left(.w_{i-1}.\right) + D_{(3;2)}N_2\left(.w_{i-1}.\right) + D_{(3;3+)}N_{3+}\left(.w_{i-1}.\right) } + \\
                        & \frac{  D_{(3;3+)}N_{3+}\left(.w_{i-1}w_{i}\right) - D_2 } { D_{(3;1)}N_1\left(.w_{i-1}.\right) + D_{(3;2)}N_2\left(.w_{i-1}.\right) + 
							D_{(3;3+)}N_{3+}\left(.w_{i-1}.\right) } + \\
                        & \gamma_{MDKNSPOMD}\left(w_{i-1}\right)p_{MDKNSPOMD}\left(w_i\right)				  
\end{aligned}
\end{equation}

with the normalization factor being:

\begin{equation}
\begin{aligned}
&\gamma_{MDKNSPOMD}\left(w_{i-1}\right) = \\ 
&\frac{D_2N_{N_{1+}}\left(.w_{i-1}.\right)}{ D_{(3;1)}N_1\left(.w_{i-1}.\right) + D_{(3;2)}N_2\left(.w_{i-1}.\right) + D_{(3;3+)}N_{3+}\left(.w_{i-1}.\right) }
\end{aligned}
\end{equation}

Finally, the unigram level can be formulated as follows:

\begin{equation}
p_{MDKNSPOMD}\left(w_i\right) = \frac{ N_{N_{1+}}\left(..w_i\right) } { N_{N_{1+}}\left(...\right) }
\end{equation}

with $N_{N_{1+}}\left(..w_i\right)$, $N_{N_{1+}}\left(.w_{i-1}.\right)$ and $N_{N_{1+}}\left(...\right)$ defined as:

\begin{equation}
\begin{aligned}
N_{N_{1+}}\left(..w_i\right) &= |\{w_{i-1} : N_{1+}\left(.w_{i-1}w_i\right) \geq 1\}| \\
N_{N_{1+}}\left(.w_{i-1}.\right) &= |\{w_i : N_{1+}\left(.w_{i-1}w_i\right) \geq 1\}| \\
N_{N_{1+}}\left(...\right) &= |\{w_{i-1},w_i : N_{1+}\left(.w_{i-1}w_i\right) \geq 1\}| \\
\end{aligned}
\end{equation}

With such a multilevel model, it is advantageous to set a different set of discount parameters at each level, based on the properties of that level (e.g. with the previously shown formulas applied to each level). However, please note that because of the non-integer values at the bigram level, it is not possible to use 3 different discounting parameters the same way as done at the trigram level, therefore only a single discount parameter ($D_2$) is used at the bigram level.

There are a couple of further details to be considered. First, negative values have to be avoided at each level, which can be achieved by discounting in the form $max(0, c-D)$ instead of simple subtraction. Moreover, in case of the $(n-1)^{th}$ level, the basic values should represent the sum of the discounts truly subtracted at the $n^{th}$ level for the given trigrams, considering the possibly reduced discount values due to the used $max$ function. To avoid over-complicated equations, these properties were not included in the above formulas.

My method can work very well for any model, even for ones with a huge number of zeros. However, I have to note that in case of the recursively interpolated version, it only preserves the marginal probabilities at the highest level (e.g. in case of a trigram model only the marginals in the form $p(..w_i)$ are preserved, and the marginals in the form $p(.w_{i-1}w_i)$ are not). Preserving the others is not possible in such a case, as there exists only one $(n-1)^{th}$ level probability distribution that preserves all the marginals, and that is exactly the one without further interpolation from it. Nevertheless, the MDKNSPOMD model still has better properties than the KNS and MKNS models in case of recursive interpolation, as neither of them keeps any of the original marginals in such a case, with the MKNS not keeping them in case of simple two-level smoothing either.

\section{Evaluation methodology and results}

To see how well my proposed MDKNSPOMD method performs compared to previous ones, I have chosen a standard n-gram language modelling evaluation task. I have used the British National Corpus (version 2; BNC, $\sim$100M words)\footnote{\url{http://www.natcorp.ox.ac.uk/}} and the text of the full English Wikipedia database dump of 01.12.2015 (EnWiki, $\sim$2000M words)\footnote{The plain text from the Wikipedia database dump was obtained with the help of the Wikipedia Extractor (\url{http://medialab.di.unipi.it/wiki/Wikipedia_Extractor}) by Giuseppe Attardi.} to evaluate the models on, both of which I previously pre-processed. Among other pre-processing steps, all text was converted to be fully lowercase. Further, in case of an n-gram model, for each sentence I added n-1 special characters at its beginning as sentence starting symbols (\textless s\textgreater, \textless s2\textgreater, etc.) and a special character at its end marking the end of the sentence (\textless /s\textgreater), to be able to fully evaluate all the meaningful words of the sentences.

In case of each corpus, I used the words occurring at least 10 times in it as vocabulary, resulting in a vocabulary size of $\sim$100k in case of the BNC, and $\sim$1.2M in case of the EnWiki corpus. Special and punctuation tokens were always considered as separate words. To achieve robust results, the average entropy and perplexity was computed using 10-fold cross-validation (inside the folds the evaluation was done sentence-by-sentence). All tests were conducted with a slightly modified version of the Kyoto Language Modelling Toolkit (Kylm)\footnote{\url{http://www.phontron.com/kylm/}}.

To be able to draw detailed conclusions, tests with both 2- and 3-gram models were conducted. In case of models above the bigram level there would be two possibilities, as noted before: to only smooth the highest level, namely only interpolating back to the second-to-highest level, or to interpolate each $n^{th}$ level with the $(n-1)^{th}$ level recursively, all the way back to the $1^{st}$ (unigram) level. However, as noted before, when the second-to-highest level is not smoothed further, then it leaves far too many zeros in a language model. This would result in zero probabilities for many sentences, making it impossible to use in practice for this type of task, and resulting in an entropy and perplexity of Infinity during evaluation. Because of this only the results for the variants that use smoothing at all levels are presented here.

All my results are shown in Table \ref{tab:table1}. These confirm previous findings that, with the use of multiple discount parameters, the MKNS slightly outperforms the original KNS in all the test cases, with both having a clear advantage over simple absolute discounting (Abs). Further, it comes as no surprise that in case of all smoothing methods, results on the 3-gram models are always remarkably better than on the 2-gram models.

Looking at the choice of the corpus, there is only a slight difference in the results in case of the 2-gram models. From this I assume that the much larger size of the EnWiki corpus is compensated by the fact that the BNC is a much more balanced corpus, containing only well-written and grammatically correct texts, and having a much narrower scope in terms of the topics covered. However, in case of the 3-grams the EnWiki corpus has a very clear dominance over the BNC, which is no surprise, as training a 3-gram model requires much more training data than a 2-gram model, so in this case clearly the relevance of the larger size of the training corpus becomes more important than the advantages of the BNC.

\begin{table}
\caption{Detailed results of the tested smoothing algorithms.}
\label{tab:table1}
\begin{center}
\resizebox{18.5pc}{!}{
\begin{tabular}{|l|l|l|r|r|}
\hline
\bf Method & \bf Model & \bf Corpus & \bf Perplexity & \bf Entropy\\
\hline
\multirow{4}{*}{Abs}  & \multirow{2}{*}{2-gram} & BNC & 252.15  & 7.96 \\ \cline{3-5}
 &   &  EnWiki & 251.82 & 7.98 \\ \cline{2-5}
 &  \multirow{2}{*}{3-gram} &  BNC & 156.01  & 7.26 \\ \cline{3-5}
 &   &  EnWiki  & 113.94 & 6.83 \\ \cline{2-5}
\hline
\multirow{4}{*}{KNS}  & \multirow{2}{*}{2-gram} & BNC & 242.57  & 7.91 \\ \cline{3-5}
 &   &  EnWiki  & 246.43 & 7.94 \\ \cline{2-5}
 &  \multirow{2}{*}{3-gram} &  BNC & 139.46  & 7.10 \\ \cline{3-5}
 &   &  EnWiki  & 104.76 & 6.71 \\ \cline{2-5}
\hline
\multirow{4}{*}{MKNS}  & \multirow{2}{*}{2-gram} & BNC & 241.18  & 7.90 \\ \cline{3-5}
 &   &  EnWiki  & 245.96 & 7.94 \\ \cline{2-5}
 &  \multirow{2}{*}{3-gram} &  BNC & 136.82  & 7.08 \\ \cline{3-5}
 &   &  EnWiki  & 103.72 & 6.69 \\ \cline{2-5}
\hline
\multirow{4}{*}{MDKNSPOMD}  & \multirow{2}{*}{2-gram} & BNC & 241.17  & 7.90 \\ \cline{3-5}
 &   &  EnWiki  & 245.98 & 7.94 \\ \cline{2-5}
 &  \multirow{2}{*}{3-gram} &  BNC & 137.50  & 7.08 \\ \cline{3-5}
 &   &  EnWiki  & 104.18 & 6.70 \\ \cline{2-5}
\hline
\end{tabular}
}
\end{center}
\end{table}

When comparing my results to that of the previous methods I can see that the MDKNSPOMD consistently outperforms the simple absolute discounting as well as the original KNS. It achieves approximately the same results as the MKNS, although there seem to be a consistent very small margin between them in favour of the MKNS in case of the 3-gram models.

Beside the evaluation in a standard language modelling task, I also plan to test my novel smoothing method in other applications too, including my methods for the automatic interpretation of noun compounds \cite{dobo2011} and the automatic computation of semantic similarity of words \cite{dobo2013}.

\section{Summary and conclusions}

Within this paper I have given a detailed overview of the motivation for smoothing methods and the most frequently used smoothing techniques. I have shown that, despite its excellent performance, the MKNS does not preserve the marginal distributions of the original data, which would be advantageous in many cases, and which, according to Goodman \cite{goodman01}, would be a requirement for a smoothing method to be optimal under certain assumptions. To overcome this problem, I have shown a modified version of the MKNS, called the MDKNSPOMD, that leaves the original marginal distributions unchanged. I have also shown that this is the only possible way of achieving this.

Beside simple problems, this model can be generalized to higher-order models too. For problems with a fairly low number of zeros it is usually enough to smooth the highest level, in which case all the original marginals are preserved. If there are too many zeros, one has to recursively smooth the lower levels too, but this way only the marginals at the highest level are preserved (it is impossible to preserve the other marginals in such a case). Nevertheless, the MDKNSPOMD is still better than the KNS and MKNS methods here too, as neither of them would keep any of the original marginals at any of the levels in such a case.

To compare this novel smoothing method with previous techniques, thorough tests have been conducted on a standard language modelling task, using two different corpora and evaluating on both 2- and 3-gram models using 10-fold cross-validation. The results show that the MDKNSPOMD performs better than both simple absolute discounting and the KNS in case of all settings, and achieves about the same results as the MKNS, with the MKNS seeming to have a minor superiority in case of the 3-gram models.

Based on this I can conclude that the novel MDKNSPOMD could be successfully used in any problem where smoothing is required, and should definitely be preferred over other methods in case preserving the marginal distributions is required or would be advantageous.

\bibliographystyle{splncs}

\begin{thebibliography}{10}

\bibitem{ney91}
Ney, H., Essen, U.:
\newblock On smoothing techniques for bigram-based natural language modelling.
\newblock In: 1991 International Conference on Acoustics, Speech, and Signal
  Processing. (1991)  825--828

\bibitem{kneser95}
Kneser, R., Ney, H.:
\newblock Improved backing-off for m-gram language modeling.
\newblock In: 1995 International Conference on Acoustics, Speech, and Signal
  Processing. (1995)  181--184

\bibitem{chen99}
Chen, S., Goodman, J.:
\newblock An empirical study of smoothing techniques for language modeling.
\newblock Computer Speech and Language \textbf{13} (1999)  359--394

\bibitem{goodman01}
Goodman, J.T.:
\newblock A bit of progress in language modeling.
\newblock Computer Speech \& Language \textbf{15} (2001)  403--434

\bibitem{teh06}
Teh, Y.W.:
\newblock A bayesian interpretation of interpolated kneser-ney.
\newblock Technical Report TRA2/06, School of Computing, National University of
  Singapore (2006)

\bibitem{siivola07}
Siivola, V., Hirsimaki, T., Virpioja, S.:
\newblock On growing and pruning kneser--ney smoothed $ n $-gram models.
\newblock IEEE Transactions on Audio, Speech, and Language Processing
  \textbf{15} (2007)  1617--1624

\bibitem{sundermeyer11}
Sundermeyer, M., Schl{\"u}ter, R., Ney, H.:
\newblock On the estimation of discount parameters for language model
  smoothing.
\newblock In: The 12th Annual Conference of the International Speech
  Communication Association. (2011)  1433--1436

\bibitem{zhang14}
Zhang, H., Chiang, D.:
\newblock Kneser-ney smoothing on expected counts.
\newblock In: The 52nd Annual Meeting of the Association for Computational
  Linguistics. (2014)  765--774

\bibitem{laplace25}
marquis~de Laplace, P.S.:
\newblock Essai philosophique sur les probabilit{\'e}s.
\newblock Bachelier (1825)

\bibitem{good53}
Good, I.J.:
\newblock The population frequencies of species and the estimation of
  population parameters.
\newblock Biometrika (1953)  237--264

\bibitem{witten91}
Witten, I.H., Bell, T.C.:
\newblock The zero-frequency problem: Estimating the probabilities of novel
  events in adaptive text compression.
\newblock IEEE Transactions on Information Theory \textbf{37} (1991)
  1085--1094

\bibitem{jelinek80}
Jelinek, F., Mercer, R.:
\newblock Interpolated estimation of markov source parameters from sparse data.
\newblock In: Workshop on Pattern Recognition in Practice. (1980)

\bibitem{chen00}
Chen, S.F., Rosenfeld, R.:
\newblock A survey of smoothing techniques for me models.
\newblock IEEE Transactions on Speech and Audio Processing \textbf{8} (2000)
  37--50

\bibitem{roark13}
Roark, B., Allauzen, C., Riley, M.:
\newblock Smoothed marginal distribution constraints for language modeling.
\newblock In: The 51nd Annual Meeting of the Association for Computational
  Linguistics. (2013)  43--52

\bibitem{dobo2011}
Dob{\'o}, A., Pulman, S.G.:
\newblock Interpreting noun compounds using paraphrases.
\newblock Procesamiento del Lenguaje Natural \textbf{46} (2011)  59--66

\bibitem{dobo2013}
Dob{\'o}, A., Csirik, J.:
\newblock Computing semantic similarity using large static corpora.
\newblock In van Emde Boas~et al., ed.: SOFSEM 2013: Theory and Practice of
  Computer Science, LNCS, vol 7741, Springer, Berlin, Heidelberg (2013)
  491--502

\end{thebibliography}

\end{document}